\documentclass[conference]{IEEEtran}
	\usepackage{filecontents,lipsum}
	
	\usepackage[noadjust]{cite}
	\usepackage{amssymb}
	\usepackage{times}
	\usepackage{epsfig}
	\usepackage{graphicx}
	\usepackage{amsmath}
	\usepackage{amssymb}
	\usepackage{gensymb}
		\usepackage[T1]{fontenc}
		\usepackage[utf8]{inputenc}

	\usepackage{varioref}
	\usepackage{array}
	\usepackage{multirow}
	\usepackage{footmisc}
	\usepackage{breakurl}
	\usepackage{mathtools}
	\usepackage{hhline}
	\usepackage{footnote}
	\usepackage{cases}
	
	\usepackage{listings}
	\usepackage{booktabs}

	\usepackage{tabularx}
	\usepackage{lipsum}
	\usepackage{capt-of}
	\usepackage{tikz}
	\usepackage{mathrsfs}
	\usetikzlibrary{arrows}
	\usepackage{epsfig}	
	\usepackage{floatrow}
	\usepackage{array}
	
	\usepackage{array}
	\newcolumntype{P}[1]{>{\centering\arraybackslash}p{#1}}
	\newcolumntype{M}[1]{>{\centering\arraybackslash}m{#1}}
	
	\usepackage{graphicx,caption,subcaption}
	\usepackage{breqn}
	\DeclareGraphicsExtensions{.pdf,.jpeg,.png,.tif}
	\makeatletter
	\def\BState{\State\hskip-\ALG@thistlm}
	\usepackage[figuresright]{rotating}	
	\usepackage{breqn}
	\usepackage{algcompatible}
	\usepackage[ruled,vlined]{algorithm2e}
	\usepackage{algpseudocode}
	
	\usepackage[pagebackref=true,breaklinks=true,letterpaper=true,colorlinks,bookmarks=false]{hyperref}
	
	\usepackage[utf8]{inputenc}

	\usepackage{amsthm}
	
	\theoremstyle{definition}

	\theoremstyle{remark}

\begin{document}
		\title{Performance Evaluation of Deep Generative Models for Generating Hand-Written Character Images}
		
		\author{\IEEEauthorblockN{Tanmoy Mondal, LE Thi Thuy Trang, Mickaël Coustaty and Jean-Marc Ogier}
			\IEEEauthorblockA{{L3i, Universit\'{e} de La-Rochelle, La-Rochelle, France}.\\
				\{tanmoy.mondal, mickael.coustaty, jean-marc.ogier\}@univ-lr.fr}
		}
		\maketitle
		
		\begin{abstract}
			There have been many work in the literature on generation of various kinds of images such as Hand-Written characters (MNIST dataset), scene images (CIFAR-10 dataset), various objects images (ImageNet dataset), road signboard images (SVHN dataset) etc. Unfortunately, there have been very limited amount of work done in the domain of document image processing. Automatic image generation can lead to the enormous increase of labeled datasets with the help of only limited amount of labeled data. Various kinds of Deep generative models can be primarily divided into two categories. First category is auto-encoder (AE) and the second one is Generative Adversarial Networks (GANs). In this paper, we have evaluated various kinds of AE as well as GANs and have compared their performances on hand-written digits dataset (MNIST) and also on historical hand-written character dataset of Indonesian BALI language. Moreover, these generated characters are recognized by using character recognition tool for calculating the statistical performance of these generated characters with respect to original character images.      
		\end{abstract}
		

		\section{Introduction}
		Today’s world of high quality document digitization has provided a stirring alternative to preserve precious ancient manuscripts. It has provided easy, hassle-free access of these ancient manuscripts for historians and researchers. Retrieving information from these knowledge resources is useful for interpreting and understanding history in various domains and for knowing our cultural as well as societal heritage. However, digitization alone cannot be very helpful until these collections of manuscripts can be indexed and made searchable. The existing characters in the document should be recognized and there are several machine learning based approaches in the literature for this purpose. But one of the primary necessity of these kinds of training based systems is the availability of labeled training dataset. Labeling datasets is not only a costly process but also highly rigorous and error prone. That's why, in this research work, we propose to automatically generate character images with the help of labeled dataset. Later these generated (that's why would be labeled automatically) dataset could be used for the training purpose which would inherently enhance the performance of the classification system.    
		
		There have been several work in the domain of Deep Generative Model (DGM) to generate  various kinds of images e.g. object images, scene images etc. In this work, we perform an evaluation of some of the popular DGMs and tested their performance on 2 datasets. The description of these datasets are given below.
		
		Principally, there are 2 main category of DGM exists in the literature. The first one is auto-encoder (AE) and the second one is Generative Adversarial Network (GAN)
		As part of this paper, we will focus on both types of the generative models. 
		
		\section{Auto-Encoder}
		An auto-encoder \cite{Ng} is an unsupervised machine learning technique, is a artificial neural network employed to recreate the given input. It takes a set of unlabeled inputs, encodes them and then tries to extract the most valuable information from them.
		Traditionally, the reduction in dimensionality is dependent on linear methods such as Principal Component Analysis (PCA), which finds the directions of maximum variance in large data. 
		However, the linearity of PCA imposes significant limitations on the types of extracted dimensional characteristics. AE overcoming these limitations by exploiting the inherent non-linearity of neural networks.
		An auto-encoder consists of three components: the coding model, code and decoding model. The purpose of the encoder function is to create a (multiple) hidden layer(s) that contains one code to describe the input. The decoder function then reconstructs the input using this code only. It has an important role during training, to force the auto-encoder to select the most important features in the compressed representation.
		
		\subsection{Vanilla auto-encoder and Multilayer auto-encoder}
		To build an auto-encoder \cite{GaleoneAutoencoder} \cite{Ng}, we need three things: a coding method, a decoding technique and a loss function to compare the output with the target. We will explore them in the next section.
		\subsubsection{Architecture}
\begin{figure}[h]
			\centering
			\includegraphics[width=9cm, height=5.5cm]{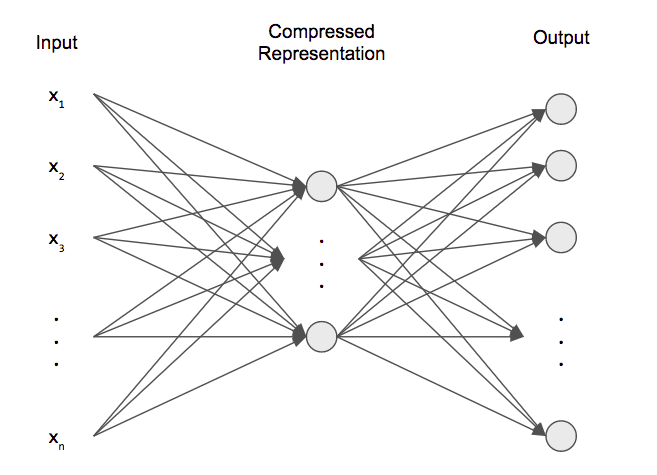}
			\caption[bla bla bla]{Vanilla auto-encoder \protect\footnotemark.}
			\label{valilla-ae}
		\end{figure}

		The encoder is an \emph{f} function that maps an entry \emph{x} to the hidden representation \emph{h}. It has the form: \(h = f (x) = s_f (W_x + b_h)\), where \(s_f\) is a nonlinear activation function, typically a logistic sigmoid \(h = f (x) = \dfrac {1} {1+ e ^ {- h}}\). The encoder is parameterized by a weight matrix \emph{W} and a bias vector \(b_h\).
		The decoder is a function\emph{g} maps the hidden representation \emph{h} return to a reconstruction \(\hat {x}\): \( \hat {x} = g (h) = s_g (W'_h + b _ { \hat {x}})\) where \(s_g\) is an activation function. The decoder parameters are the weight matrix \({W '} \) and a bias vector \(b _ {\ hat {x}}\).
		In its simplest form, auto-encoder is a two-layer network, i. e. a fully connected feed-forward neural network with hidden layer(s). The architecture of vanilla auto-encoder is shown in Figure \ref{valilla-ae}, whose input and output layer have the same number of neurons, the hidden layer is smaller than the size of the input and output layer. The hidden layer is a compressed representation, and we learn two sets of weights and bias that encode our input data in the compressed representation and decode our compressed representation in the input space.
		
		\footnotetext{Source: \url{https://pythonmachinelearning.pro/all-about-autoencoders/}}	
		A natural thought that may arise is to extend auto-encoder beyond the single layer; which can be easily achieved just by keeping the dimensionality of entry and exit same, where as increasing the number of hidden layers.
		\paragraph{Loss function}
		An auto-encoder tries to learn an approximation of the identity function, to produce the reconstruction \( \hat {x} \) which is similar to the input \(x \). The loss function is calculated either by using the mean squared error \emph{MSE} or by binary cross entropy.
		If the input values are in the range[0-1], then we typically use the cross entropy loss function.
		\begin{align}
		L(f(x)) = -\sum_{i=1}x_i log(\hat{x}) + (1 - x_i) log(1 - \hat{x})
		\label{Equation-auto_encoder_loss-1}
		\end{align}
		Otherwise, we have a simple mean squared loss:
		\begin{align}
		L(f(x)) = \dfrac{1}{2}\sum_{i=1}(x_i - \hat{x}_i)^2
		\label{Equation-auto_encoder_loss-2}
		\end{align}
		Our goal is to minimize this loss function. This error represents how close our reconstruction is to the true input data. We don't expect a perfect reconstruction because the number of hidden neurons is less than the number of input neurons, but we want the parameters to give us the best possible reconstruction.
		
		\subsection{Sparse auto-encoder}
		In the vanilla auto-encoder, we assume that the number of hidden units is small. But even when the number of hidden units is greater than the number of input units, we can still discover some interesting representation of the input data. To achieve this for a given input, most of the hidden neurons should produce only weak activation \cite{Ng}.
		In other words, its average activation value should be a small  (sigmoid activation function gives activation value close to 0 and \(\tanh \) activation function gives activation value close to -1). In this sort of specific structure, the auto-encoder will discover an interesting structure in the data. Which inherently means that, for a given instance, only an informative set of units is activated, so that more discriminating representation could be captured. 
		
		The average activation of each hidden unit: \(\hat {\rho_j} = = \dfrac{1}{m}\sum_{i=1}^m [a_j(x^i)]\),  where  \(a_j\) indicates activation of this hidden unit. The constraint is imposed by \(\rho = \hat {\rho_j}\), where \(\rho\) is \emph{"sparsity parameter"}.
		To achieve this, we have to add an additional penalty term to our optimization objective which penalizes \(\hat {\rho_j}\) significantly deviating from \(\rho\): \( \sum_{j=1}^{S_2}KL(\rho ||\hat {\rho_j})
		\), where \(S_2\) is the number of units in the hidden layer, $j$ is the index of the hidden unit in the network, KL divergence is a standard function to measure the difference between two different distributions: \(KL(\rho ||\hat {\rho_j}) = \rho\log\dfrac{\rho}{\rho_j} + (1-\rho) \log\dfrac{1-\rho}{1-\hat{\rho_j}}\)
		If \(\rho = \hat {\rho_j}\), this penalty function has the property that \(KL(\rho ||\hat {\rho_j}) = 0\). Otherwise, it increases monotonously when \(\hat{\rho}\) diverge from \( \rho\).

		\ Our overall cost function is now becomes:
	  \begin{equation}
		  \begin{multlined}
			\label{sparce_autoencoder_2}
			J_(sparse) = L(f(x)) + \beta\sum_{j=1}^{S_2}KL(\rho ||\hat {\rho_j})
		  \end{multlined}
	 \end{equation}
		\ where \(L(f(x))\) is defined in the vanilla auto-encoder; \(\beta\) control the weight of the term sparsity parameter.
		\subsection{Convolutional auto-encoder}
		So far, we have seen that the auto-encoder inputs are images. It is therefore logical to ask whether a convolutional architecture can work better than the classical auto-encoder architectures previously discussed. Instead of using fully connected layers, we use convolution and grouping layers to reduce our input to a coded representation \cite{convolutionalAE-Galeone}.
		We recall that the auto-encoder consists of two parts: coding and decoding. For coding, we use a traditional convolutional neural network whose main mechanism for reducing information in this convolutional network is the \emph{max-pooling} layer.
		To resize our encoded representation to the same form as the encoding, a simpler operation is used to increase the spatial size of the representation. Unlike the \emph{max-pooling} technique, \emph{un-pooling} technique is used. This layer corresponds to the inverse of the max-pooling operation under certain simplifying conditions. The \emph{un-pooling} layer is performed by simply replacing each entry of a feature map with a \(s \times s \) block with the input value in the top left corner and zeros elsewhere.
		
		\subsection {De-Noising auto-encoder}
		A de-noising auto-encoder is an extension of the convolutional auto-encoder. Suppose we have an input image with noise (these noisy images are actually pretty common in real-world scenarios). For a de-noising auto-encoder \cite{denoisingAE_2}, the model we use is identical to the convolutional auto-encoder. However, our training and test data are different. For our training data, we add random Gaussian noise, and our test data is the original and clean images. Our input data is the: \(x' = x + \alpha * \epsilon\), where $\alpha$ is a percentage of the amount of noise applied to the input images and \(\epsilon \sim N(0, \sigma^2I)\) is the distribution for generating Gaussian noise.           
		This causes the de-noising auto-encoder to produce clean images from noisy images given as input to the system.
		
		\subsection {Contractive auto-encoder - \emph{CAE}}
		The aim of a contractive auto-encoder is to make the learned representation be robust towards small changes around its training examples. 
		The contractive auto-encoder \cite{Rifai2011}, is a special form of regulated auto-encoder that is trained to minimize the following regularized reconstruction error:
		\begin{align}
		\label{4}
		L = \sum_{x \in \mathfrak {D}} (L(x,g(f(x))) + \lambda |J_f(x)|_F ^2)
		\end{align}
		where \(|J_f(x)|_F ^2 = \dfrac{\delta h_i(x)}{\delta x_i}\). $\mathfrak {D}$ represents the complete training dataset. $F$ is the Frobenius norm and \(\lambda\) is the positive parameter that control the regularization. 
		Note that the success of the minimization of CAE criterion strongly depends on the parameter \(f\) and \(g\) and in particular the tied weight constraint used, with \(f(x) = s_f(W_x + b_h)\) and $g(h) = s_g(W'_h + b_{\hat{x}})$. Where $h$ represents hidden/encoded representation obtained from given input $x$ and $\hat x$ represents the generated output obtained from $h$. The above regularization term forces \(f\) (as well as \(g\), because of the related weights) to be contractive, that is to have singular values lower than $1$. The higher values of \(\lambda\) give more contraction (smaller singular values) but in the local directions where there is little or no variations of data, the degree of data contraction is less.
				
		\subsection {Variational auto-encoder - \emph{VAE}}
		In the language of neural networks, a variational auto-encoder \cite{Doersch2016} consists of: A probabilistic encoder \(Q(h|x)\) and a generative decoder \(P(\hat x|h)\) and a loss function. Where $h$ represents hidden/encoded representation obtained from given input $x$ and $\hat x$ represents the generated output obtained from $h$.  The  weights and biases for encoder is mentioned by $\theta$ and $\phi$ for decoder. 
		In the decoding process, information is lost because it passes from a smaller to a larger dimension. The amount of information lost must therefore be measured using the reconstruction \emph{log-likelihood}: \(\log P(\hat x|h)\). This measure signifies how effectively the decoder has learned to reconstruct an input image $x$ given its latent representation $h$.
		The \emph{loss function} of the variational auto-encoder is the negative log-likelihood with a regularizer. The following defined loss function is decomposed into only terms which depends on single data point $l_i$. The total loss then becomes $\sum_{i=1}^{N}l_i$ for total $N$ data points. The loss function $l_i$ for data point $x_i$ is
		  \begin{equation}
		  \begin{multlined}
			  \label{variational_autoencoder}
			  	l_i(\theta, \phi) = - E_{h \sim Q_\theta(h|x_i)} \left[\log P(x_i|h)\right] - KL \left[ Q_\theta(h|) || P(h)\right]
		  \end{multlined}
		  \end{equation}
		The $1^{st}$ term in Equation \ref{variational_autoencoder} is the reconstruction loss or expected negative log-likelihood of the $i^{th}$ data point and the $2^{nd}$ term is the \emph{Kullback-Leibler} divergence between the encoder's distribution $q_\theta(h|x)$ and $p(h)$, which measures information loss (in units of nats) when using $q$ to represent $p$.

		\subsection {Conditional Variational auto-encoder - \emph{CVAE}}
		The conditional variational auto-encoder is an extension of the variational auto-encoder. The VAE aims to formulate the problem of data generation as a Bayesian model. This model is learned by optimizing its lower limit. However, we have no control over the VAE data generation process. This could be problematic if we want to generate specific data. That is why the CVAE has been developed. While VAE models mainly latent variables and data directly, CVAE models latent variables and  data \cite{conditionalVariationalAE}, both conditioned by few random variables.
		For CVAE, the model is now conditioned to two variables $x$ and \emph{c}: The encoder \(Q(h|x,c)\); the decoder \(P(\hat x|h,c)\). So, the goal we take is to:
		\begin{align}
		L = E_{h \sim Q} \left[\log P(\hat x|h, c)\right] - D_{KL} \left[ Q(h|x, c) || P(h|c)\right]
		\end{align}
		We have just conditioned all distributions with a variable \emph{c}. Now, the latent variable is distributed under \(P(h|c)\).

		\section {Variations of \emph{GANs}}
		\subsection{Generative Adversarial Nets }
		\subsubsection{Architecture}
		The GAN \cite{Goodfellow} estimates a generative model via a contradictory process by simultaneously forming two models: The generator - \emph {G}, which creates samples intended to come from the same distribution as the learning data; The discriminator - \emph {D} learns using traditional techniques of supervised learning, dividing the entries into two classes (real or false). The architecture of GANs is shown in Figure \ref{GANs_Model} below.
		\begin{figure}[h]
			\centering
			\includegraphics[trim = 0.0cm 0.8cm 0.4cm 2.9cm, clip,  scale=0.45]{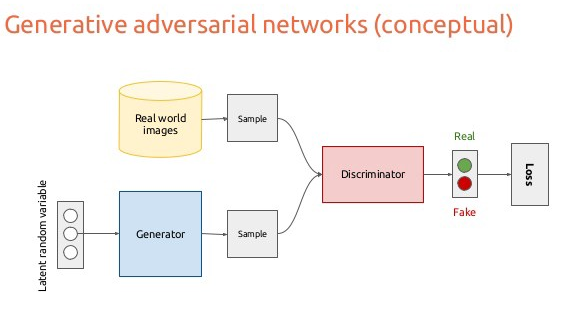}
			\caption[bla bla bla]{Generative Adversarial Networks \protect\footnotemark.}
			\label{GANs_Model}
		\end{figure}
		\footnotetext{Source: \url{https://wiki.tum.de/pages/viewpage.action?pageId=23562510}}	
		\subsubsection{The learning process}
		For the generator, we start by sampling the $z$ vector of the distribution a priori. The generator function ($G$) is applied to the input vector $z$. The generator function is a differential function with parameters that can be learned with gradient descent. 
		The discriminator (type of differential function) is the opposite of generator which is fed by the generated images and by certain training images at the same time. It is learned by descending gradient similar to the generative function. The goal of generator is to generate image looks like to the real ones whereas the discriminator's goal is to discriminate real ones from generated ones.
		
		\subsubsection{The loss function}
		$D$ take $x$ as input and use \(\theta^{(D)}\) as parameters, whereas \emph{G} take $z$ as input and use\(\theta^{(G)}\) as parameters. Both players have loss functions that are defined according to these parameters. 
		The discriminator wants to minimize \(J^D(\theta^{(D)} \theta^{(G)}\) and must do so by controlling only \(\theta^{(D)})\). 
		The generator wants to minimize \(J^G(\theta^{(D)} \theta^{(G)})\) and must do so while controlling only \(\theta^{(G)}\). 
		The loss function for the discriminator which is presented below:
		\begin{multline}
		J^D(\theta^{(D)} \theta^{(G)} = \\ E_{x\sim p_{data}(x)} [logD(x)] + E_{z\sim p_{z}(z)} [log(1 - D(G(z))    
		\end{multline}
		This is just the standard cost of cross entropy that is minimized when forming a standard binary classifier with a sigmoid output. 
		\(J ^ {(G)} \) is directly related to \(J ^ {(D)} \), we can summarize the whole game with a value function specifying the gain of the discriminant:
		\begin{align}
		V(\theta^{(D)}, \theta^{(G)}) = -J^{(D)}(\theta^{(D)}, \theta^{(G)})
		\end{align}
		Zero-sum games are also called minimax games because their solution involves minimizing in an external loop and maximizing in an internal loop.
		\begin{multline}
		\min_{G}\max_{D}V(D,G) = \\
		E_{x\sim p_{data}(x)} [logD(x)] + E_{z\sim p_{z}(z)} [log(1 - D(G(z))]  
		\end{multline}
		
		\subsection{\emph{Conditional Generative Adversarial Networks - CGANs}}
		M. Mirza and S. Osindero \cite{Mirza2014a} extend the GAN model by conditioning both networks $D$ and $G$ by an additional $c$ parameter, which could be any type of auxiliary information, such as class labels or data from other modalities. In this context, the value function is changed as follows:
		\begin{multline}
		\min_{G}\max_{D}V(D,G) = \\
		E_{x\sim p_{data}(x)} [logD(x|c)] + E_{z\sim p_{z}(z)} [log(1 - D(G(z|c))] 
		\end{multline}
		The CGANs are interesting for two reasons: First, the CGANs learn how to use the additional information and therefore, they are able to generate better samples. Secondly, with CGANS, we have a way of controlling image representations. For example, in the case of face generation, with GANs, all information is encoded by $z$. With CGANs, when we add conditional information to it, these two $z$ and $c$ now encode different information $c$ could describe attributes such as hair color, skin color or gender.
		
		\subsection{\emph{Deep Convolutional Generative Adversarial Networks - DCGAN}}
		A.Radford \cite{Radford2015b} presents a topologically constrained variant of the conditional $GAN$. To build a $DCGAN$, two deep convolutional neural networks are used. The first network consists of deep architecture which is used to look at a picture and processes it through several layers to recognize increasingly complex features in the image. Whereas, the second neural network is learned to create false images. $DCGANs$ propose modifications to $GANs$ by
		replacing all layers of \emph{pooling} with stride convolutions (for discriminant) and fractional stride convolutions (for generators). Batch normalization is used in the generator (all layers except the output layer) and in the discriminator (all layers except the input layer). Leaky $ReLU$ is used in all layers of the discriminator and $ReLU$ activation is used in all layers of the generator (except the output layer that uses $Tanh$ activation function).
		
		\subsection{\emph{Wasserstein Generative Adversarial Networks - WGAN}}
		
		\ In fact, in the training procedure based on GANs, two models (each model updates its cost independently) are trained simultaneously to find a balance between two-players non-cooperative game. Therefore, it is unknown when to stop training (no convergence).
		The classical GAN's minimize the divergence of \emph{Jensen-Shanon} which is equal to zero if the actual and false distribution does not overlap (which is the usual case). Thus, instead of minimizing \emph{Jensen-Shanon} divergence, we can use \emph{Wasserstein's} distance ($W$). 
		WGAN \cite{Arjovsky2017} adds some tricks to allow the discriminant to approach the \emph{Wasserstein} distance between the real distributions and models. The authors propose to approach \(W \) with a set of functions \emph {K-Lipschitz} by solving the following problems:
		
		\begin{align}
		\max_{w \in W} E_{x\approx P_r} - E_{z \approx p(z)} [f_w (G_{\theta} (z))]
		\end{align}
		
		\ The distance from Wasserstein is also called Earth Mover’s distance - EMD, which is defined by following Equation \ref{WasserDist}.
		
		\begin{align}
		W(P_r, P_\theta) = \inf_{\lambda \in \prod(P_r, P_\theta)}E_{(x,y)\inf \gamma}[|x-y|] 
		\label{WasserDist}
		\end{align}
		
		\ where \(\prod(P_r, P_\theta) \) means all the joint distributions \(\lambda(x, y)\) whose marginals are respectively \(P_r\) et \(P_g\)
		

		
		\  The authors argue that compared to vanilla GAN, WGAN has the following advantages: Significant Loss Measure: The loss of D correlates well with the quality of the generated samples, allowing less monitoring of the training process; Improved stability: When D is trained to the optimum, it provides a useful loss for G training. This means that the training of D and G must not be balanced in number of samples (it must be balanced in the vanilla GAN approach). 
		\subsection{Adversarial Auto- encoder - \emph{AAE}}
		\ One of the main disadvantages of variation auto-encoders is that the integral of the \emph{KL-divergence} term has no closed form analytical solution except for a handful of distributions. Moreover, it is not easy to use discrete distributions for the latent $z$ code. Indeed, back-propagation by discrete variables is generally not possible, which makes the model difficult to train effectively. AAE \cite{Makhzani2015a} is an approach to do so in the context of the VAE has been introduced.
		AAE avoids using \emph{KL-divergence}  altogether by using contradictory learning. In this architecture, a new network is formed to discriminatingly predict whether a sample is from the hidden code of the auto-encoder or priority distribution \(p (z) \) determined by the user. 

		\begin{figure}[h]
			\centering
			\includegraphics[width=9cm, height=5.5cm]{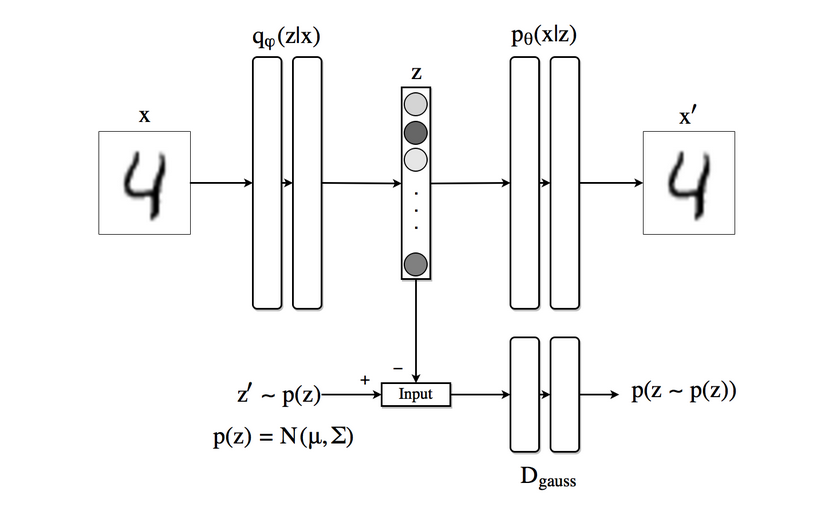}
			\caption[bla bla bla]{Adversarial auto-encoder \protect\footnotemark.}
				\label{AAE_image}
		\end{figure}
		\footnotetext{Source: \url{https://blog.paperspace.com/adversarial-autoencoders-with-pytorch/}}	
		\ Figure \ref{AAE_image} shows schematically how AAE works when we use a Gaussian a priori for the latent code (although the approach is generic and can use any distribution). The top row is equivalent to a VAE. First, a sample \(z \) is plotted against the generator network \(Q (z | X) \), this sample is then sent to the decoder that generates \(x '\) from \(z \). The reconstruction loss is computed between \(x \) and \(x '\) and the gradient is retro-propagated by \(P \) and \(Q \) accordingly.
		\section{Dataset and Experimental Protocol}
		\subsection{Dataset}
		In this work, we have used following 2 datasets.
		\subsubsection{MNIST Dataset}
		It is handwritten dataset consists of $60,000$ handwritten digit (0-9) images for training and $10,000$ images for testing. The size of these handwritten digit images is normalized and the digits are centered in a fixed-size image to fit into a $28 \times 28$ pixel space in binary format.
		\subsubsection{BALI Dataset}
		This is BALInese palm manuscript images dataset comes from BALI, Indonesia. The sample images are randomly selected from 23 different collection (contents) with total of 393 pages. The isolated character dataset is formed by segmenting all the patch images and annotating them at the character level. It consists of total $133$ character classes, with total number of $19, 383$ character samples. Among $133$ classes $50$ classes are chosen because it contains more than $300$ to $500$ samples. From each class $250$ images are chosen for training purpose and then remaining $20$ images are chosen for the testing purpose.
		
		\subsection{Experimental Protocol}
		Each of the aforementioned models are tested for several times with different parameters such as the learning rate, batch size and number-of-epochs and the obtained best results of each model are presented here.  For all the models, the batch-size is taken as $128$ and the learning rate is taken as $0.001$ but the number-of-epoch varies for every model. Different models are tested on 2 datasets by using then following parameters. For AE, CAE, convolutional AE, de-noising AE, AAE, GAN, cGAN, dcGAN; number-of-epoch is taken as $100000$ and for SPAE, VAE, CVAE, wGAN; the number-of-epoch is taken as $50000$.
		For models such as AE, CAE, SPAE, convolution-AE, de-noising AE, we used testing images as the input to test model performance. For the remaining models, we do not use the testing database because the models takes distribution a priori as the input. However, for models such as cGAN and CVAE, image labels are also provided as the input to generate the desired output class. The generated images by different models are shown in the following Table \ref{allModelsImages}.
		
		To evaluate the performance of the system, the generated images are recognized by using character recognition algorithm. The recognition of generated characters are only applied for those models which can generate images of a predefined class; i.e. AE, CAE, SPAE, convolutional-AE, de-noising-AE, CVAE, cGAN. However by simple visual inspection it can be visible that the quality of images generated by the convolution auto-encoder and de-noising auto-encoder models are not good. We therefore apply the character recognition technique (defined below) only for the following models : AE, CAE, SPAE,CVAE, cGAN. 
		\subsubsection{Brief Description of Recognition Technique}
		\ We use \emph{Convolutional Neural Network}  based recognition system to recognize the generated images 
		\ For MNIST database, the following CNN architecture is used. This network consists of at first a convolutional layer which take $64$ convolution matrices of size $5 \times 5 + LeakyReLu$ followed by \emph{Max-Pooling} of size $2 \times 2$, and strides = $(2 \times 2)$. 
		The second convolutional layer takes $32$ convolution matrices of size $5\times5 + LeakyReLu$ followed by a \emph{Max-Pooling} layer of $2 \times 2$ with strides = $(2 \times 2)$. Which results in dimension reduction from $7 \times 7 \times 32$ to $1 \times 1 \times 1024$ which is then fed into fully connected neural network is applied to classify the images into $10$ classes.
		
		In case of BALI database, deeper CNN based architecture is used due to the bad image quality of BALI database.  
		The architecture of this CNN is as follows: The first convolutional layer take $64$ convolution matrices of size $3 \times 3 + LeakyReLu$, followed by \emph{Max-Pooling} layer of size $2 \times 2$, strides = $(2 \times 2)$. 
		The second convolutional layer takes $32$ convolution matrices of size $3 \times 3 + LeakyReLu$, followed by a \emph{MaxPooling} layer of size $2 \times 2$, strides = $(2 \times 2)$.
		 The third convolutional layer takes $32$  convolution matrices of size $3 \times 3 + LeakyReLu$, followed by a \emph{MaxPooling} layer of size $2 \times 2$, strides = $(2 \times 2)$. 
		 The feature vector is reduced from $4 \times 4 \times 32$ dimensions to $1 \times 1 \times 100$ dimensions which is then fed into fully connected neural network to classify the images into $40$ classes.
		
		\begin{table}[h]
			\begin{center}
				\begin{tabular}{c c c} 
					\textbf{Model name} & \textbf{MNIST} & \textbf{BALI}\\
					\hline
					AE & 89.98/97.25 & 40.01/60.01\\
					CAE & 96.79/97.25 & 47.5/60.01\\
					SPAE & 95.92/97.25 & 54.99/60.01\\
					CVAE & 92.40/97.25 & 35.01/60.01\\
					cGAN & 87.52/97.25 & 29.97/60.01
				\end{tabular}
			\end{center}
			\caption{Accuracy on Generated/ Original Images}
			\label{Recognition_Result}
		\end{table}
		
		\newcolumntype{C}{>{\centering\arraybackslash}m{9.2em}}
		\newcolumntype{D}{>{\centering\arraybackslash}m{13em}}
		
		\begin{table*}[ht]
			\begin{tabular}{l*5{C}@{}}
				& \includegraphics[width=9.2em]{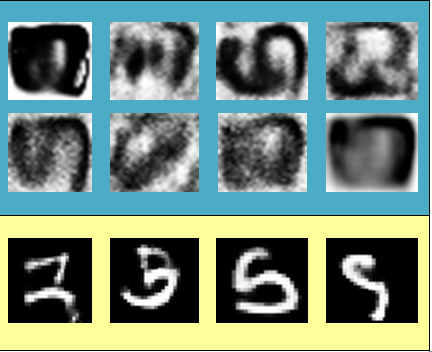}
				& \includegraphics[width=9.2em]{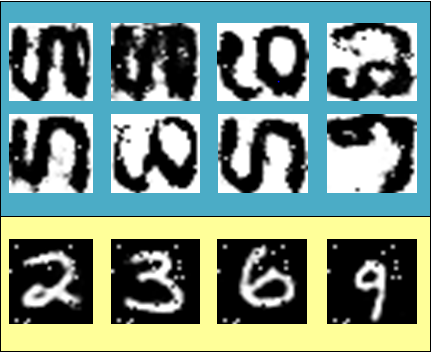}
				& \includegraphics[width=9.2em]{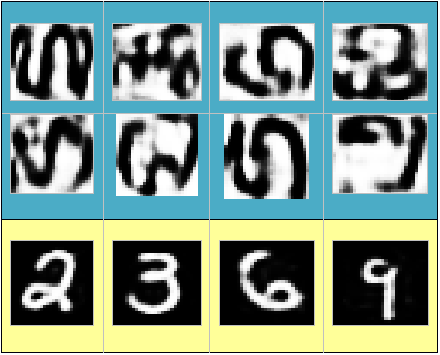}
				& \includegraphics[width=9.2em]{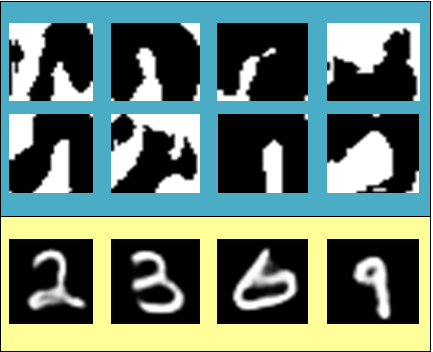} 
				& \includegraphics[width=9.2em]{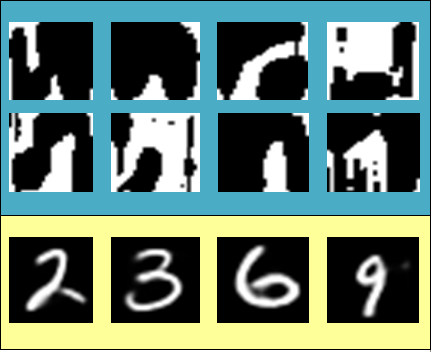} \\ 
				& (a) & (b) & (c) & (d) & (e)\\ 
			\end{tabular}
			
			\begin{tabular}{l*3{D}@{}}
				& \includegraphics[width=13em]{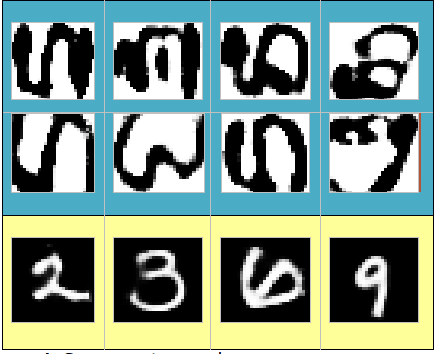}
				& \includegraphics[width=13em]{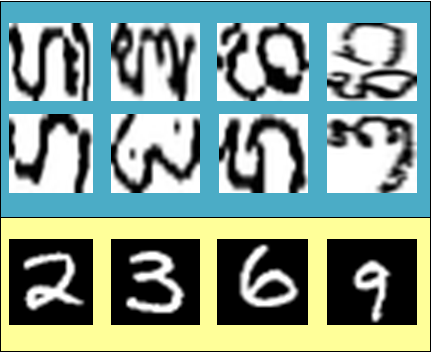}
				& \includegraphics[width=13em]{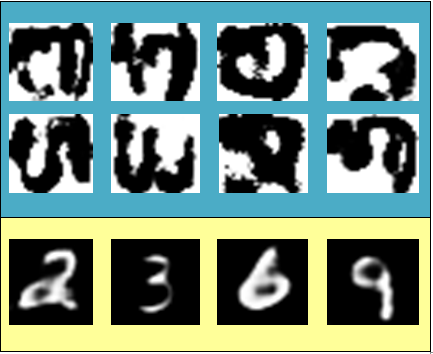} \\
				& (f) & (g) & (h) \\ 
			\end{tabular}
			
			\begin{tabular}{l*5{C}@{}}
				& \includegraphics[width=9.2em]{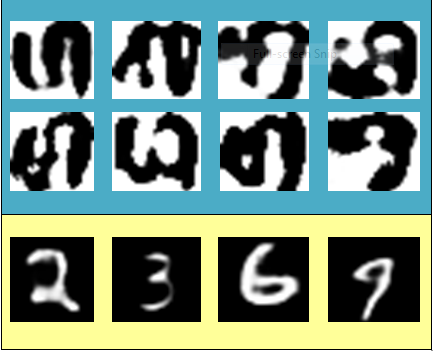}
				& \includegraphics[width=9.2em]{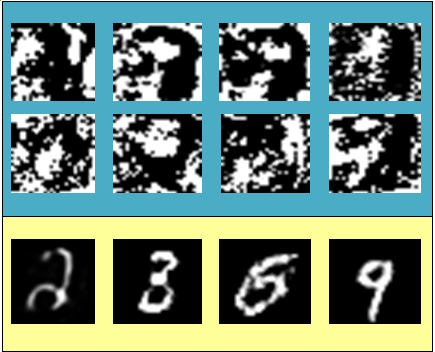}
				& \includegraphics[width=9.2em]{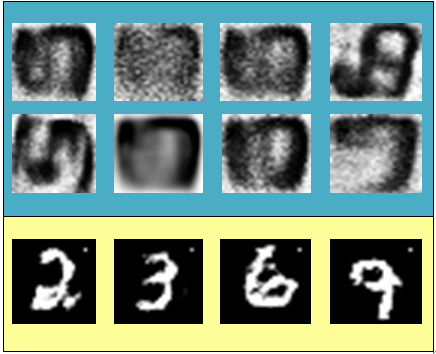}
				& \includegraphics[width=9.2em]{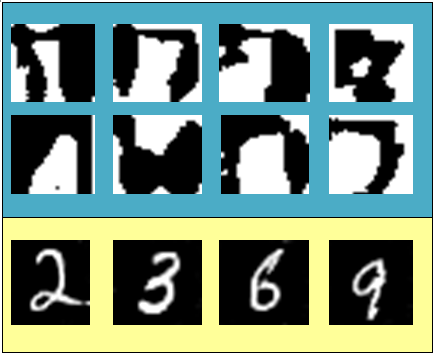} 
				& \includegraphics[width=9.2em]{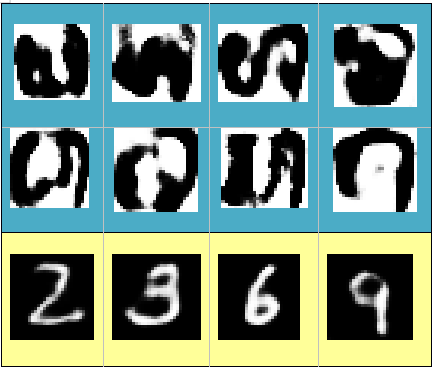} \\ 
			\centering
			& (i) & (j) & (k) & (l) & (m) \\ 
			
			\end{tabular}
			\caption{The image generated by various models are shown below. The images generated for BALI dataset is contained in the blue colored region and the images for MNIST dataset is contained in the yellow colored region of each of the following box. Where each following box represents the examples of generated images by various models. (a): Results of Adversarial Auto-Encoder   (b): Results of Auto-Encoder  (c): Results of Contractive Auto-Encoder  (d): Results of  Convolutional Auto-Encoder  (e): Results of De-noising Auto-Encoder
			(f):Results of Sparse Auto-Encoder (g): The original images (h): Results of Variational Auto-Encoder    
			(i): Results of Contractive Variational Auto-Encoder  (j): Results of GAN  (k): Results of Conditional GAN   (l): Results of Deep-Convolutional GAN  (m):Results of Wasserstien GAN }
			\label{allModelsImages}
		\end{table*}
		\section{Results and Discussion}
		The recognition results are shown in \ref{Recognition_Result}. It can be visible that CAE, SPAE and CVAE has performed better in the case of MNIST dataset whereas SPAE has performed well for BALI dataset.
		In case of BALI dataset, most of the auto-encoder based models function better than GANs based models. Because, the main idea of auto-encoder based models is to reconstruct the  original images from the hidden representations. While, the GANs based models try to generate the images from the a priori distribution. This is the reason why GANs need several learning images compared to auto-encoder. Moreover, the quality of the images in BALI database are degraded and noisy. So, it is difficult for generative models to reconstruct such images. 
		Among all the auto-encoders \cite{Rifai2011}, the SPAE model gives the best results because this model adds an additional penalty term to the optimization function. This term allows the SPAE to learn representation robust towards small changes around its training examples. 
		The convolutional neural network needs many samples for each class (does not works well with BALI dataset), as the one exists for MNIST dataset ($ \approx 60000$).
		Among the GANs based models, the wGAN works better than others because wGAN resolves the convergence problem (exists in classical GANs model) during learning process by using the Wasserstein distance. 
		
		\section{Conclusion and Future Work}
		In this article, we tested the various models of two kinds of generative model (GAN and Auto-encoder) with two data sets: MNIST and BALI. From the experimental evaluation, it is visible that certain models work well with BALI and MNIST data sets and certain are not. 
		In the future, we plan to work with GANs based model to improve the performance of BALI dataset by proposing certain techniques to define better the apiori distribution as input for generating any particular class of images (i.e. improving cGAN model). 
		
		\section*{Acknowledgment}

		\bibliographystyle{IEEEtran}
		\bibliography{IEEEabrv,library}  
	\end{document}